\documentclass[letterpaper, 10 pt, conference]{ieeeconf}  

\IEEEoverridecommandlockouts                              

\usepackage{graphics} 
\usepackage{amsmath} 
\usepackage{amssymb}  
\usepackage[font={small,it}]{caption}
\usepackage{subcaption}
\usepackage{algorithm}
\usepackage{algorithmic}
\usepackage{dblfloatfix}

\usepackage{tikz}
\usetikzlibrary{positioning, shapes.geometric, arrows}

\tikzstyle{state}=[thick,draw=black,circle,minimum size = 7mm,inner sep=0mm]
\tikzstyle{action}=[draw=black,fill=black,circle,inner sep=0mm, minimum size=1.5mm]
\tikzstyle{arrow} = [thick,->]
\tikzstyle{process} = [rectangle, minimum width=0.9cm, minimum height=0.9cm, text centered, draw=black]

\title{\LARGE \bf
Reinforced Grounded Action Transformation for Sim-to-Real Transfer
}

\author{Haresh Karnan$^{1\S}$, Siddharth Desai$^{1\S}$, Josiah P. Hanna$^{2}$, Garrett Warnell$^{3}$ and Peter Stone$^{4}$
\thanks{$^\S$Equal contribution}%
\thanks{$^{1}$ The University of Texas at Austin, Department of Mechanical Engineering
        {\tt\footnotesize \{sidrdesai,haresh.miriyala\}@utexas.edu}}%
\thanks{$^{2}$School of Informatics, University of Edinburgh; To be joining the Computer Sciences Department, University of Wisconsin---Madison
        {\tt\footnotesize josiah.hanna@ed.ac.uk}}%
\thanks{$^{3}$Army Research Laboratory
        {\tt\footnotesize garrett.a.warnell.civ@mail.mil}}%
\thanks{$^{4}$The University of Texas at Austin, Department of Computer Science and Sony AI
        {\tt\footnotesize pstone@cs.utexas.edu}}%
}
\begin{document}
\maketitle
\thispagestyle{empty}
\pagestyle{empty}
\begin{abstract}

Robots can learn to do complex tasks in simulation, but often, learned behaviors fail to transfer well to the real world due to simulator imperfections (the ``reality gap").
Some existing solutions to this sim-to-real problem, such as Grounded Action Transformation (\textsc{gat}), use a small amount of real-world experience to minimize the reality gap by ``grounding" the simulator. 
While very effective in certain scenarios, \textsc{gat} is not robust on problems that use complex function approximation techniques to model a policy.
In this paper, we introduce {\em Reinforced Grounded Action Transformation} (\textsc{rgat}), a new sim-to-real technique that uses Reinforcement Learning (RL) not only to update the target policy in simulation, but also to perform the grounding step itself.
This novel formulation allows for end-to-end training during the grounding step, which, compared to \textsc{gat}, produces a better grounded simulator.
Moreover, we show experimentally in several MuJoCo domains that our approach leads to successful transfer for policies modeled using neural networks.

\end{abstract}

\section{INTRODUCTION}

In reinforcement learning (RL), the {\em sim-to-real problem} entails effectively transferring behaviors learned in simulation to the real world. Often, learning directly on the real world can be too time-consuming, costly, or dangerous. Using a simulator mitigates these issues, but simulators are often imperfect models, leading to learned policies that are suboptimal or unstable in the real world. In the worst cases, the simulated agent learns a policy that exploits an inaccuracy in the simulator---a policy that may be very different from a viable real world solution.

A promising paradigm for addressing the sim-to-real problem is that of Grounded Simulation Learning (\textsc{GSL}) \cite{AAMAS13-Farchy}, in which one seeks to modify (i.e., {\em ground}) the simulator to better match the real world based on data from the real world.
If the internal parameters of the simulator cannot be easily modified (as is often the case in practice), the state-of-the-art grounding approach is Grounded Action Transformation (\textsc{gat}) \cite{hanna2017grounded}.
\textsc{Gat} performs grounding not by modifying the simulator, but rather by augmenting it with a learned {\em action transformer} that seeks to induce simulator transitions that more closely match the real world.
Hanna and Stone demonstrate that \textsc{gat} can transfer a bipedal walk from a simulator to a physical NAO robot. The complex dynamics involved with a multi-actuated robot walking on soft carpet make it very difficult to create an accurate simulator for the domain. Whereas training in simulation without \textsc{gat} produces a highly unstable real-world policy, the parameters learned with \textsc{gat} produced the fastest known stable walk on the NAO robot \cite{hanna2017grounded}.

In parallel to development in the sim-to-real space, there has been an explosion of interest in using deep neural networks to represent RL policies. Successes of Deep RL include milestone achievements such as mastering the game of Go \cite{alphagozero} and solving a Rubik's cube with one robotic hand \cite{openairubikscube}. In the robotic motion domains that we consider in this work, deep learning is a key component of most leading RL algorithms such as Trust Region Policy Optimization (\textsc{trpo}) \cite{trpoalgo}, Proximal Policy Optimization (\textsc{ppo}) \cite{ppoalgo}, and Soft Actor Critic (\textsc{sac}) \cite{sacalgo}.

Unfortunately, trying to combine deep RL with sim-to-real grounding approaches has proven difficult, which limits the policy representations possible for sim-to-real problems. In \textsc{gat} \cite{hanna2017grounded}, the policy learned was optimized over sixteen parameters. The number of parameters of a neural network are many orders of magnitude higher. We find that trying to use \textsc{gat} with neural network policies often fails to produce transferable policies (see Section \ref{sim2real}). We hypothesize that {\em this poor performance is due to imprecision in the grounding step} and that {\em learning the action transformer end-to-end can improve transfer effectiveness}. 

To test this hypothesis, we introduce {\em Reinforced Grounded Action Transformation} (\textsc{rgat}), a new algorithm that modifies the network architecture and training process of the action transformer.
We find that this new grounding algorithm produces a more precise action transformer than \textsc{gat} with the same amount of real-world data.
We perform simulation experiments on OpenAI Gym MuJoCo domains, using a modified simulator as a surrogate for the real world. Using this surrogate  allows us to compare our sim-to-real approach with training directly on the ``real" world, which is often not possible on real robots.
\footnote{Of course, doing so comes with the risk that the methods developed may not generalize to the real world. In this paper, we focus on developing a novel training methodology for learning in simulation. Conducting extensive evaluation of this approach is only possible with a surrogate real world. Other work \cite{hanna2017grounded, sgat} has focused on evaluating similar methods on real robots, and such experimentation with \textsc{rgat} is an important direction for future work.}
We find that \textsc{rgat} outperforms \textsc{gat} at transferring policies from sim to ``real" when using policies represented as deep neural networks, and matches the performance of an agent that is trained directly on the ``real'' environment, thus confirming our hypothesis.

\section{BACKGROUND}

Motivated by increasing interest in employing data-intensive RL techniques on real robots, the sim-to-real problem has recently received a great deal of attention.
Sim-to-real is an instance of the transfer learning problem.
As we define it, sim-to-real refers to transfer between domains where the transition dynamics differ and the rewards are the same.
Note that with this formalism, it is not strictly necessary for the sim domain to be virtual nor for the real domain to be physical.\footnote{Indeed in transfer learning terminology, the sim and real domains would be called source and target domains respectively. In this paper, we will primarily use ``sim" and ``real."}
This section summarizes the existing sim-to-real literature and specific literature from reinforcement learning related to our proposed approach.

\subsection{Related Work}

The sim-to-real literature can be broadly divided into two categories of approaches. Methods in the first category seek to learn policies robust to changes in the environment. In applications where the target domain is unknown or non-stationary, these methods can be very useful. Dynamics Randomization adds noise to the environment dynamics, which has led to success in finding robust policies for robotic manipulation tasks \cite{DBLP:journals/corr/abs-1710-06537}. Action Noise Envelope (\textsc{ane}) \cite{Jakobi1995} randomizes the environment by adding noise to the action. While these methods uses noise injected at random to modify the environment, Robust Adversarial Reinforcement Learning (\textsc{rarl}) uses an adversarial agent to modify the environment dynamics \cite{DBLP:journals/corr/PintoDSG17}. Using a different paradigm, meta-learning attempts to find a meta-policy which can be learned in simulation and then can quickly learn an actual policy on the real environment \cite{DBLP:journals/corr/FinnAL17}.

Methods in the second category, which we refer to as grounding methods, seek to improve the accuracy of the simulator with respect to the real-world. Unlike the robustness methods, these methods have a particular target real-world domain and usually require collecting data from it. We can think of grounding methods as strategies to correct for simulator bias, whereas the robustness methods only correct for simulator variance. System identification type approaches try to learn the exact physical parameters of the system---either through careful experimentation as done with the Minitaur robot \cite{DBLP:journals/corr/abs-1804-10332} or through more automated methods of system identification like TuneNet \cite{DBLP:journals/corr/abs-1907-11200}. Often, these methods require alternating between improving the simulator and improving the policy as in Grounded Simulation Learning \cite{AAMAS13-Farchy}. Our approach follows this basic format, but unlike these methods (and like \textsc{gat} \cite{hanna2017grounded}), we do not assume that we have a parameterized simulator that we can modify. Neural-Augmented Simulation (\textsc{nas}) \cite{pmlr-v87-golemo18a} and policy adjustment \cite{higuera17} are similar approaches to \textsc{gat} but use different neural architectures and training procedures.

\textsc{gat} achieved remarkable success on a challenging domain; however, there has not been much work applying it to different domains. Our approach improves upon the \textsc{gat} algorithm to overcome some of its limitations.

\subsection{Preliminaries}

Formally, we treat the sim-to-real problem as a reinforcement learning problem \cite{rlsutton}. The real environment is a Markov Decision Process (\textsc{mdp}). At each time step, $t$, the environment's state is described by $s_t \in \mathcal{S}$. The agent samples an action, $a_t \in \mathcal{A}$, from its policy, $a_t \sim \pi(\cdot|s_t)$. The environment then produces a next state: $s_{t+1} \sim T_{real}(\cdot|s_t,a_t)$, where $T_{real}$ is the transition probability distribution. The agent also receives a reward, $r_{t+1} \in \mathbb{R}$, from a {\em known} function of the action taken and the next state: $r_{t+1} = R(a_t, s_{t+1})$. In the controls literature, this is often called a cost function (which is a negative reward function). The discount factor $\gamma \in [0,1]$ controls the relative utility of near-term and long-term rewards. The RL problem is to find a policy, $\pi$, that maximizes the expected sum of discounted rewards: $\Sigma_{t=0}^{\infty} \gamma^t R(a_t, s_{t+1})$

The simulator is an \textsc{mdp} that differs only in the transition probabilities, $T_{sim}$. The sim-to-real objective is to maximize the expected return for the RL problem while minimizing the number of time steps evaluated on the real \textsc{mdp}. The tradeoff between these objectives depends on the specific application.

\subsection{Grounded Action Transformation (\textsc{gat})}

\begin{figure}[tb]
\centering
\begin{tikzpicture}[node distance=2cm]
\node (pi) [process, label=Policy, fill=blue!20] {$\pi_\theta$};
\node (at) [process, below of=pi, xshift=1.2cm, yshift=-0.5cm, minimum width=3.3cm, minimum height=1.3cm, fill=red!20, label=below:{Action Transformer}] {};
\node (inv) [process, below of=pi, xshift=0.25cm, yshift=-0.5cm, fill=white] {$f^{-1}_{sim}$};
\node (fwd) [process, right of=inv, xshift=-0.1cm, fill=white] {$f_{real}$};
\node (sim) [process, left of=inv, fill=green!20] {Sim};
\draw[arrow] (fwd) -- node [below] {$\hat{s}_{t+1}$} (inv);
\draw[arrow] (inv) -- node [below] {$\hat{a}_t$} (sim);
\coordinate [right of=fwd, xshift=-1cm] (cr);
\draw[arrow] (pi) -| node[above]{$a_t$} (cr) -- (fwd);
\coordinate [left of=sim, xshift=1cm] (cl);
\draw[arrow] (sim) -- node [below, xshift=-0.2cm, yshift=0.1cm] {$s_{t+1}$} (cl) |- node [above] {$s_t$} (pi);
\coordinate [left of=pi, yshift=-0.8cm, xshift=-1cm] (dl);
\coordinate [right of=pi, yshift=-0.8cm, xshift=1.7cm, label={Agent}, label=below:{Env}] (dr);
\draw[dashed] (dl) -- (dr);
\node (loss) [process, above of=sim, yshift=-1cm] {$R$};
\coordinate [above of=loss, yshift=-1cm, label={$r_t$}] (rew);
\draw[arrow] (loss) |- (rew);
\draw[arrow] (cr) |- (loss);
\draw[arrow] (cl) |- (loss);
\end{tikzpicture}
\caption{Diagram of the \textsc{gat} training process \cite{hanna2017grounded}, showing how the forward model, $f_{real}$, and the inverse model, $f^{-1}_{sim}$, transform the action, $a_t$, before it passes to the simulator. Everything below the agent/env boundary is considered the grounded simulator.}
\label{fig:GAT}
\end{figure}
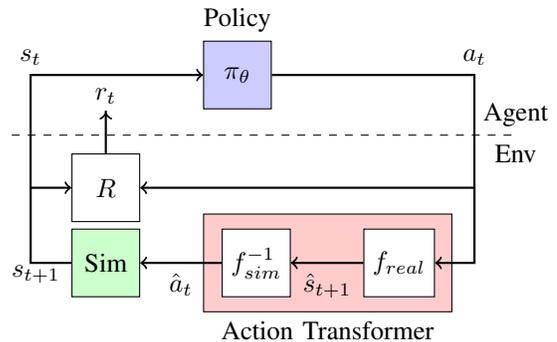

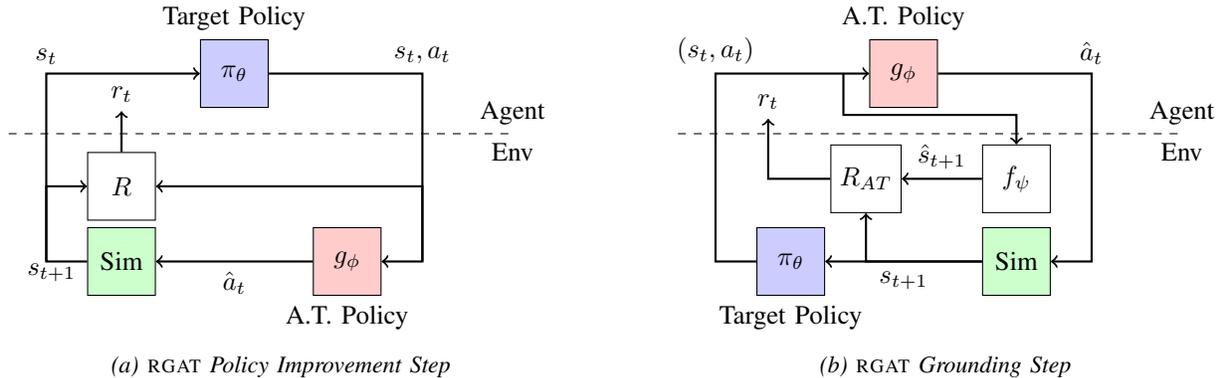
\begin{figure*}[!hb]
\begin{subfigure}{.49\textwidth}
\centering
\begin{tikzpicture}[node distance=2cm]
\node (pi) [process, label={Target Policy}, fill=blue!20] {$\pi_\theta$};
\node (at) [process, below of=pi, xshift=1.5cm, yshift=-0.5cm, label=below:{A.T. Policy}, fill=red!20] {$g_\phi$};
\node (sim) [process, left of=at, xshift=-1cm, fill=green!20] {Sim};
\draw[arrow] (at) -- node [below] {$\hat{a}_t$} (sim);
\coordinate [right of=at, xshift=-1cm] (cr);
\draw[arrow] (pi) -| node[above]{$s_t, a_t$} (cr) -- (at);
\coordinate [left of=sim, xshift=1cm] (cl);
\draw[arrow] (sim) -- node [below, xshift=-0.2cm, yshift=0.1cm] {$s_{t+1}$} (cl) |- node [above] {$s_t$} (pi);
\coordinate [left of=pi, yshift=-0.8cm, xshift=-1cm] (dl);
\coordinate [right of=pi, yshift=-0.8cm, xshift=1.7cm, label={Agent}, label=below:{Env}] (dr);
\draw[dashed] (dl) -- (dr);
\node (loss) [process, above of=sim, yshift=-1cm] {$R$};
\coordinate [above of=loss, yshift=-1cm, label={$r_t$}] (rew);
\draw[arrow] (loss) -- (rew);
\draw[arrow] (cr) |- (loss);
\draw[arrow] (cl) |- (loss);
\end{tikzpicture}
\caption{\textsc{rgat} Policy Improvement Step}
\label{fig:RGATtp}
\end{subfigure}
\begin{subfigure}{.49\textwidth}
\centering
\begin{tikzpicture}[node distance=2cm]
\node (at) [process, label={A.T. Policy}, fill=red!20] {$g_\phi$};
\node (sim) [process, below of=at, xshift=1.5cm, yshift=-0.5cm, fill=green!20] {Sim};
\node (pi) [process, left of=sim, xshift=-1cm, label=below:{Target Policy}, fill=blue!20] {$\pi_\theta$};
\draw[arrow] (sim) -- node [below] {$s_{t+1}$} (pi);
\coordinate [right of=sim, xshift=-1cm] (cr);
\draw[arrow] (at) -| node[above]{$\hat{a}_t$} (cr) -- (sim);
\coordinate [left of=pi, xshift=1cm] (cl);
\coordinate [left of=at, xshift=1.2cm] (al);
\draw[arrow] (pi) -- (cl) |- node [above] {$(s_t, a_t)$} (al) -- (at);
\node (loss) [process, below of=at, yshift=0.6cm, xshift=-0.5cm] {$R_{AT}$};
\node (fwd) [process, right of=loss] {$f_\psi$};
\coordinate [above of=fwd, yshift=-1.15cm] (fwda);
\draw[arrow] (al) |- (fwda) -- (fwd);
\draw[arrow] (fwd) -- node [above] {$\hat{s}_{t+1}$} (loss);
\draw[arrow] (sim) -| (loss);
\coordinate [above of=loss, xshift=-1.3cm, yshift=-1.2cm, label={$r_t$}] (rew);
\draw[arrow] (loss) -| (rew);
\coordinate [left of=at, yshift=-0.8cm, xshift=-1cm] (dl);
\coordinate [right of=at, yshift=-0.8cm, xshift=1.7cm, label={Agent}, label=below:{Env}] (dr);
\draw[dashed] (dl) -- (dr);
\end{tikzpicture}
\caption{\textsc{rgat} Grounding Step}
\label{fig:RGATatp}
\end{subfigure}
\caption{Diagram of the two steps of the proposed \textsc{rgat} algorithm. Note that the outer loop is the same in both steps, but only the policy on the Agent side is updated during the Policy Improvement and Grounding step.}
\label{fig:RGATdiagram}
\end{figure*}

The GSL framework \cite{AAMAS13-Farchy} consists of alternating between two steps, called the \textit{grounding step} and the \textit{policy improvement step}. During the grounding step, the target policy, $\pi$, remains frozen while the simulator is improved, and, during the policy improvement step, the grounded simulator is fixed, making this step a standard RL problem. The policy improvement step is done entirely in the grounded simulator. 
GSL continues alternating between these steps until the policy performs well on the real environment.

\textsc{gat} \cite{hanna2017grounded} introduced a particular way of grounding the simulator which treats the simulator as a black box. The grounding step for \textsc{gat} is as follows: \begin{enumerate}
    \item Evaluate the current policy on both environments and store trajectories, $\{s_0, a_0, s_1, a_1, ...\}$ as $\tau_{real}$ and $\tau_{sim}$. 
    \item Using supervised learning, train a forward model of the dynamics, $f_{real}: \mathcal{S} \times \mathcal{A} \rightarrow \mathcal{S}'$, from the data in $\tau_{real}$. This model---usually a neural network---learns a mapping from $(s_t, a_t)$ to the maximum likelihood estimate of the next state observation, $\hat{s}_{t+1}$. 
    \item Similarly, train an inverse model, $f^{-1}_{sim}: \mathcal{S} \times \mathcal{S}' \rightarrow \mathcal{A}$ from $\tau_{sim}$. This model is a mapping from two states, $(s_t, s_{t+1})$, to the action, $\hat{a}_t$, that is most likely to produce this transition in the simulator.
    \item Compose the forward and inverse models to form the {\em action transformer}, $g(s_t, a_t) = f^{-1}_{sim}(s_t, f_{real}(s_t, a_t))$.
\end{enumerate}

During the policy improvement step, the reward is still computed explicitly as $R(a_t, s_{t+1})$. A block diagram of the grounded simulator for \textsc{gat} is shown in Fig. \ref{fig:GAT}. When the action transformer is prepended to the simulator, the resulting {\em grounded simulator} produces a next state, $s_{t+1}$, that is closer to the next state observed in the real world. Thus, if we learn a policy on a good grounded simulator, the policy will also perform well on the real world.

\section{REINFORCED GROUNDED ACTION TRANSFORMATION (\textsc{rgat})}

\begin{algorithm}[!tb]
\caption{Reinforced Grounded Action Transformation}
\label{alg:RGAT}
\textbf{Input}: initial parameters $\theta$, $\phi$, and $\psi$ for target policy $\pi_\theta$, action transformer policy $g_\phi$, and forward dynamics model $f_\psi$; policy improvement methods, \texttt{optimize1} and \texttt{optimize2} 
\begin{algorithmic}[1]
\WHILE{policy $\pi_\theta$ improves on real}
 \STATE Collect real world trajectories \\$\tau_{real} \leftarrow \{((s_0,a_0),s_1), ((s_1, a_1), s_2),...\}_{real}$
 \STATE Train forward dynamics function $f_\psi$ with $\tau_{real}$
 \STATE Update $g_\phi$ in simulation by using \texttt{optimize1} and reward, $r_t = -||f_\psi(s_t,a_t) - s_{t+1}||^2$
 \STATE Update $\pi_\theta$ in simulation using \texttt{optimize2} and the reward from the grounded simulator
\ENDWHILE
\end{algorithmic}
\end{algorithm}

In our experiments, we find that \textsc{gat} produces a very noisy action transformer (see Section \ref{sim2self}). We hypothesize that this noise is due to the composition of two different learned functions---since the output of $f_{real}$ is the input to $f^{-1}_{sim}$, errors in $f_{real}$ are compounded with the errors in $f^{-1}_{sim}$.
To reduce these errors, we introduce Reinforced Grounded Action Transformation (\textsc{rgat}), an algorithm that trains the action transformer end-to-end.
Since there is no straightforward supervisory signal that can be used to train the action transformer, we propose to learn the action transformer using reinforcement learning. In \textsc{rgat}, we learn a single action transformation function for $g$ as opposed to learning $f_{real}$ and $f^{-1}_{sim}$ separately. Training the action transformer as a single neural network also allows us to learn the \textit{change} in action, $\Delta a_t=\hat{a}_t-a_t$, rather than the transformed action directly. If the simulator is realistic, then the values of $\Delta a$ will be close to $0$ indicating no change is required; however, the values for $\hat{a}$ span the whole action space. Thus, this change has a normalizing effect on the output space of the neural network, which makes training easier.

Our experiments show that \textsc{rgat} produces more precise action transformers than \textsc{gat} while using the same amount of real world data. In this approach, we treat the grounding step as a separate RL problem. Like \textsc{gat}, \textsc{rgat} first uses supervised learning to train a forward model, $f_\psi$, parameterized by $\psi$; however, unlike \textsc{gat}, $f_\psi$ is not part of the action transformer. This forward model gives a prediction of the next state $f_\psi(s_t, a_t) = \hat{s}_{t+1}$, which is used to compute the \textit{reward} for the action transformer.

Here, we model the action transformer as an RL agent with policy $g_\phi$ with parameters $\phi$. We will call this the {\em action transformer policy} to distinguish it from the {\em target policy}, the policy of the behavior learning agent we wish to deploy on the real world. This agent observes the state, $s_t$, and the action taken by the target policy, $a_t=\pi_\theta(s_t)$. Therefore, the input space for the action transformer policy is the product of the state and action spaces of the target policy $\mathcal{S}_{AT} = \mathcal{S} \times \mathcal{A}$. The output of the action transformer policy is a transformed action, so the action space remains the same $\mathcal{A}_{AT}=\mathcal{A}$. Since there are two different RL agents with different objectives, they have different reward functions. The reward for the target policy is provided by the grounded simulator whereas the reward for the action transformer policy is determined by the closeness of the transition in the grounded simulator to the real world. At each time step, the actual next state from the grounded simulator is compared to the next state predicted by the forward model and the action transformer is penalized for the difference with the per step reward $R_{AT}(\hat{s}_{t+1}, s_{t+1}) = -||\hat{s}_{t+1} - s_{t+1}||^2 = -||f_\psi(s_t, a_t) - s_{t+1}||^2$.

That is, the reward is the negative L2 norm squared between expected next state and actual next state. The difference between training the two different policies is shown in Fig. \ref{fig:RGATdiagram}. Note that the outer loop in both are exactly the same. The blocks are just rearranged to make the agent--environment boundary clear. Fig. \ref{fig:RGATatp} also shows the forward model that is used to compute $R_{AT}$. This block is missing from Fig. \ref{fig:RGATtp} since the target policy's reward is provided by the grounded simulator.

\section{EXPERIMENTS}
We designed experiments to test our hypotheses that training the action transformer end-to-end improves the precision of the action transformer and that this improved precision improves sim-to-real transfer.
First we compare the precision of the \textsc{gat} and \textsc{rgat} action transformations by examining how individual actions are transformed on the \textit{InvertedPendulum} domain (Section \ref{sim2self}).
We then evaluate the policies learned using each algorithm on the ``real'' domain to compare how well the polciies transfer(Section \ref{sim2real}). We use a modified simulator to act as a surrogate for the real world.
For these experiments, we use the MuJoCo continuous control robotic domains provided by OpenAI Gym \cite{openai-gym}.
We evaluate \textsc{rgat} on three different MuJoCo environments---\textit{InvertedPendulum-v2}, \textit{Hopper-v2} and \textit{HalfCheetah-v2}.
InvertedPendulum is a simple environment with a four dimensional continuous state space and a one dimensional continuous action space.
Both Hopper and HalfCheetah are relatively complex environments with high-dimensional state and action spaces compared to InvertedPendulum, and their dynamics are more complex due to presence of friction and contact forces.
We use an implementation of \textsc{trpo} from the stable-baselines library \cite{stable-baselines} for both \texttt{optimize1} and \texttt{optimize2} of Algorithm \ref{alg:RGAT}.

\subsection{Sim-to-Self Experiments}
\label{sim2self}

\begin{figure}[!tb]
    \centering
    \includegraphics[width=\columnwidth]{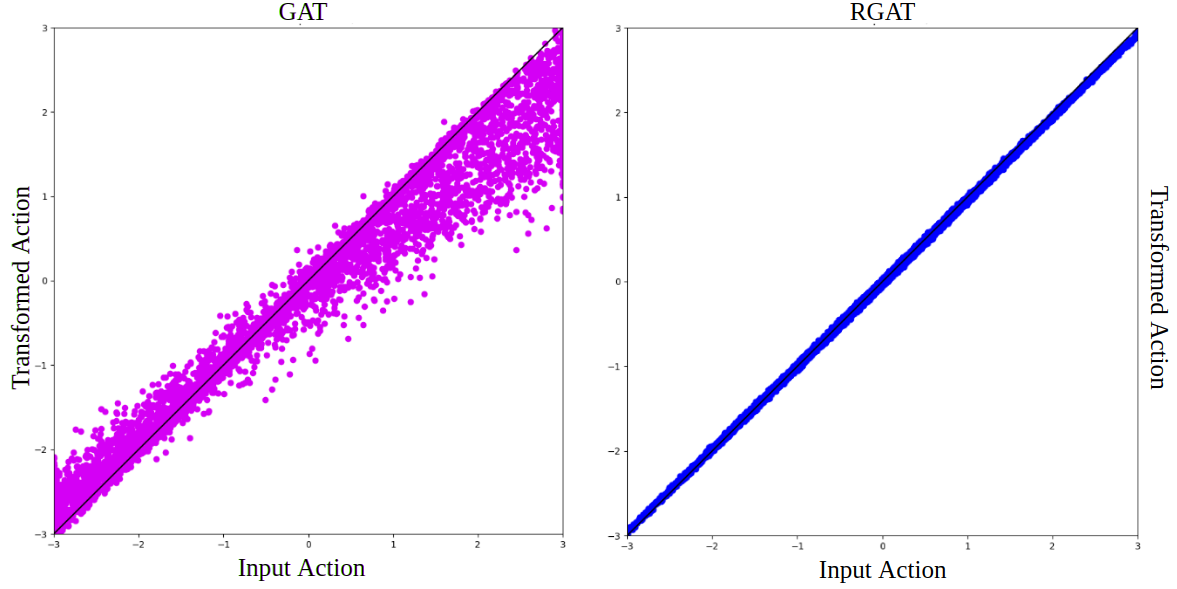}
    \caption{Transformed action vs original action in a \textbf{sim-to-self} experiment on the InvertedPendulum environment---learned using \textsc{gat} (left) and \textsc{rgat} (right) algorithms. The black line shows the fixed points of the action transformer. \textsc{rgat} has much lower variance than \textsc{gat}.}
    \label{fig:gatrgat_sim2sim}
\end{figure}

\begin{figure}[!tb]
    \centering
    \includegraphics[width=\columnwidth]{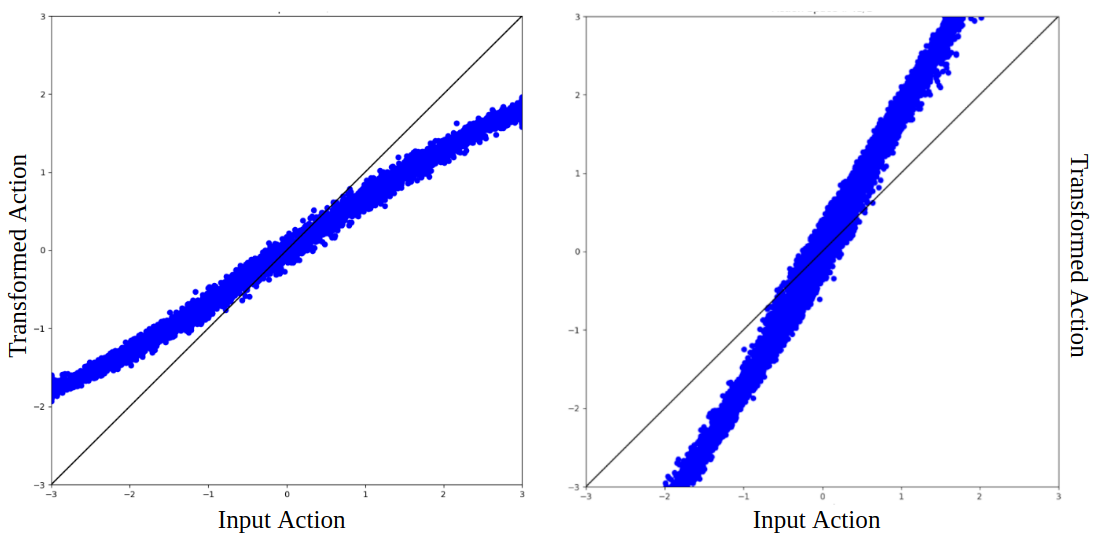}
    \caption{Transformed action vs original action in a \textbf{sim-to-real} experiment on the InvertedPendulum environment---learned using \textsc{rgat}, where the real pendulum is heavier (left) or lighter (right) than the simulated pendulum. The black line shows the fixed points of the action transformer.}
    \label{fig:sim2real_actions}
\end{figure}

\begin{figure*}[!tb]
\begin{subfigure}{\columnwidth}
    \includegraphics[width=\columnwidth]{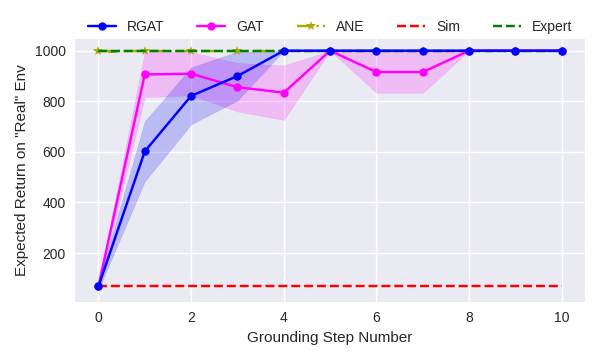}
    \caption{Shallow Network}
    \label{fig:reduced_ip}
\end{subfigure}
\begin{subfigure}{\columnwidth}
    
    \includegraphics[width=\columnwidth]{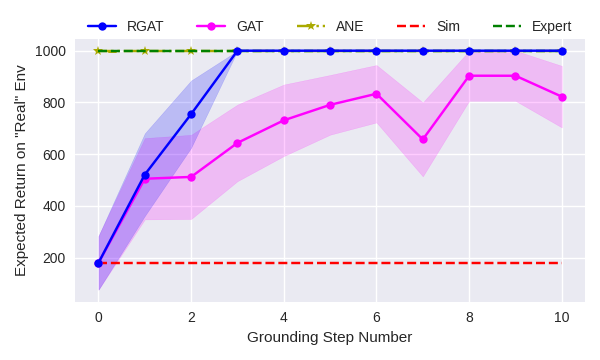}
    \caption{Deep Network}
    \label{fig:ip_main}
\end{subfigure}
\label{fig:ip}
\caption{Average performance of \textsc{rgat} and \textsc{gat} over ten grounding steps for \textbf{InvertedPendulum}. The real pendulum has a mass of 100 units. The shaded region shows standard error over ten independent training runs. In the shallow network case (a), both algorithms do well, but in the deep network case (b), \textsc{gat} fails to reach optimal performance.}
\end{figure*}

\begin{figure}[!tb]
    \centering
    \includegraphics[width=\columnwidth]{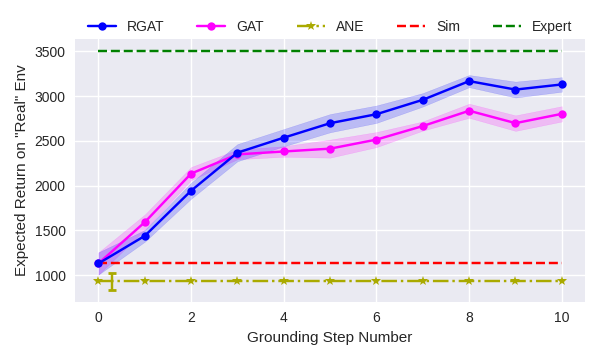}
    \caption{Average performance of \textsc{rgat} and \textsc{gat} over twenty grounding steps for \textbf{HalfCheetah}. The ``real" HalfCheetah's torso is 15\% heavier than the sim HalfCheetah. The shaded region shows standard error over ten independent training runs. \textsc{rgat} outperforms \textsc{gat}, but both algorithms eventually reach the optimal reward.}
    \label{fig:hcheetah_main}
\end{figure}

To test the precision of an action transformer, we first apply both the \textsc{gat} and \textsc{rgat} algorithms to settings where the sim and real domains were exactly the same. Ideally, during the grounding step, the action transformer should learn not to transform the actions at all. This effect is easy to visualize for InvertedPendulum, since the action space is one dimensional. Fig. \ref{fig:gatrgat_sim2sim} shows the transformed action versus the input action after one grounding step for \textsc{gat} and \textsc{rgat}. The black line shows the points where the transformed action is the same as in the input action. From the figure, we can see that \textsc{rgat} produces a better action transformer, since the dots lie much closer to the black line. The transformer for \textsc{gat} has a wider distribution with a bias toward the black line.

\subsection{Policy Representation}

Consistent with Hanna and Stone \cite{hanna2017grounded}, we find that \textsc{gat} works well on transferring policies where the policy representation is low dimensional. When we use a shallow neural network for the target policy---one hidden layer of four neurons---\textsc{gat} and \textsc{rgat} have very similar performance. We run ten trials of both algorithms, evaluating the performance on the ``real" environment after each policy improvement step. Fig. \ref{fig:reduced_ip} shows the mean return over ten grounding steps for both algorithms. For reference, we compare the results to a policy trained only in simulation (red line), and a policy that is allowed to train directly on ``real'' until convergence (green line).

We then repeat that experiment using a deeper network---a fully connected neural network with two hidden layers of 64 neurons. The sim-to-real experiment results on InvertedPendulum is shown in Fig. \ref{fig:ip_main}. \textsc{gat} fails to transfer a policy from sim to ``real", as was discussed in the previous sections. However, \textsc{rgat} receives close to the optimal reward even with a high-dimensional policy representation.

\subsection{Sim-to-``Real" Experiments}
\label{sim2real}

Similar to the action transformation visualizations shown in Section \ref{sim2self}, we can visualize the transformations for the sim-to-real case. Fig. \ref{fig:sim2real_actions} shows the action transformation graphs for two different \textit{InvertedPendulum} ``real" world environments. On the left, the ``real'' pendulum has a greater pendulum mass than the simulated pendulum. Therefore, the magnitude of the actions decreases---a weaker force on the lighter pendulum has the same effect as a stronger force on the heavier pendulum. If the real pendulum is lighter, the opposite happens, as is shown in the figure on the right.

Note that the action transformer takes both the state and action as input, so the same input action could be transformed to different output action depending on the state. Thus, this effect accounts for some of the variance in Fig. \ref{fig:sim2real_actions}, whereas in Fig. \ref{fig:gatrgat_sim2sim} the variance is only due to modeling error.

To further test the effectiveness of \textsc{rgat}, we repeat the experiment from Fig. \ref{fig:ip_main} on the HalfCheetah and Hopper domains. The target policy architecture is the same as in Fig. \ref{fig:ip_main}. For these domains, using shallower networks is not an option, because lower capacity networks fail to learn good policies, even when trained directly on the real domain.

We chose the mass for the ``real" environments based on the analysis from the \textsc{rarl} paper \cite{DBLP:journals/corr/PintoDSG17}. Changing the physical parameters of the robot results in different transition dynamics, which acts as our surrogate for the ``real" environment; however in certain cases, it can make the task much easier or harder.  We thus verify that an agent trained directly on the modified environment reaches the same optimal return as is expected on the original domain. Therefore, if a policy performs poorly on the modified simulator, we know this is because of poor transfer and not because the task is harder.

\begin{figure}[!tb]
    \centering
    \includegraphics[width=\columnwidth]{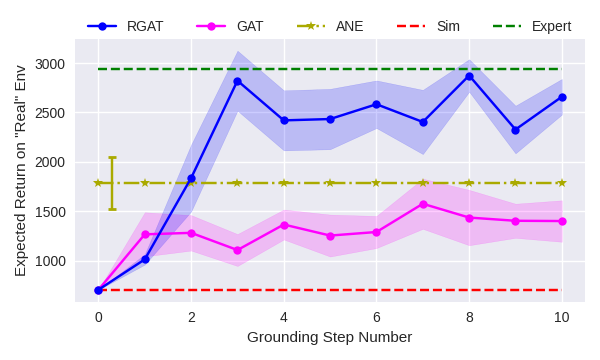}
    \caption{Average performance of \textsc{rgat} and \textsc{gat} over ten grounding steps for \textbf{Hopper}. The ``real" Hopper's torso is 27\% heavier than the sim Hopper. The shaded region shows standard error over ten independent training runs. Here, \textsc{gat} barely improves upon the baseline. \textsc{rgat} quickly reaches the green line in three grounding steps.
    }
    \label{fig:hopper_main}
\end{figure}


Figs. \ref{fig:reduced_ip}, \ref{fig:ip_main}, \ref{fig:hcheetah_main}, and \ref{fig:hopper_main} show plots comparing the performance of \textsc{gat} and \textsc{rgat}. Both algorithms use 50 real world trajectories for every grounding step. In all of these experiments, \textsc{rgat} performs significantly better than \textsc{gat} and performs as well as a policy trained directly on the "real" domain (green line). For comparison, the green lines on these plots show the performance of a policy trained directly on the real environment for up to ten million timesteps of experience.

We also compare against the Action Noise Envelope method \cite{Jakobi1995}. Like \textsc{gat} and \textsc{rgat}, \textsc{ane} is a sim-to-real method that treats the simulator as a black-box. However, \textsc{ane} is not a grounding method---it seeks only to find policies that are robust to a prespecified noise distribution. For a given target domain, this distribution is typically unknown, but we performed experiments for several specified distributions and report only the best results. In the Inverted Pendulum domain, \textsc{ane} does well, but in the more complex domains, it is unable to learn a policy that transfers well. This behavior is expected because robustness methods try to perform well over a variety of environments at the cost of performance on any one particular domain.

\subsection{Hyperparameters}
Though we use the stable-baselines library hyperparameters for policy improvement, using \textsc{trpo} as the grounding step optimizer introduces a new set of hyperparameters for the algorithm. The parameters we found to be most critical to the success of the algorithm were the \textit{maximum KL divergence} constraint and the \textit{entropy coefficient}. We found that if the action transformer policy changed too much during a single grounding step, then the target policy often failed to learn. Thus, the \textit{maximum KL divergence} should be small, but not so small that the policy cannot change at all. The \textit{entropy coefficient} should be large enough to ensure exploration. In our experiments, we set the \textit{max KL divergence} constraint value to 1e-4 and \textit{entropy coefficient} to 1e-5.

The discount factor for the action transformer, $\gamma_{AT}$, is an additional hyperparamter we can control. Since the action transformer in \textsc{rgat} is an RL agent, it may pick suboptimal actions at the present step to get a higher reward in the future. In this sense, the action transformer tries to match the whole trajectory instead of just individual transitions. Setting $\gamma_{AT}=1$ leads to matching the entire trajectory, and setting $\gamma_{AT}=0$ causes the learner to only look at individual transitions. In our experiments, we set $\gamma_{AT}$ to 0.99.

\section{DISCUSSION AND FUTURE WORK}

The experiments reported above confirm our hypotheses that learning the action transformer end-to-end improves its precision (Section \ref{sim2self}) and that policies learned using \textsc{rgat} transfer better  to the ``real" world 
than policies learned using \textsc{gat} (Section \ref{sim2real}). When the target policy network is shallow, the difference between the algorithms is less noticeable, but when the network capacity increases, inaccuracies in the action transformer have a greater effect. 



Having demonstrated success in transferring between simulators and having analyzed in detail the scenarios in which \textsc{rgat} outperforms \textsc{gat}, the next important step in this research is to validate \textsc{rgat} on physical robots.

\section{CONCLUSION}

This paper introduced Reinforced Grounded Action Transformation (\textbf{\textsc{rgat}}), a novel  algorithm for grounded simulation learning. We investigate why \textsc{gat} fails to learn a good action transformer and improve upon \textsc{gat} by learning an action transformer end-to-end. \textsc{rgat} is able to learn a policy for grounding a simulator, using limited amount of experience from the target domain, and our method is compatible with existing deep RL algorithms, such as \textsc{trpo}. We experimentally validated \textsc{rgat}'s sim-to-real performance on the InvertedPendulum, Hopper and HalfCheetah environments from MuJoCo, and we showed empirically that within a few grounding steps, \textsc{rgat} can produce a policy that performs as well as a policy trained directly on the target domain.

\section{ACKNOWLEDGMENT}
\small{This work has taken place in the Learning Agents Research Group (LARG) at UT Austin.  LARG research is supported in part by NSF (CPS-1739964, IIS-1724157, NRI-1925082), ONR (N00014-18-2243), FLI (RFP2-000), ARL, DARPA, Lockheed Martin, GM, and Bosch.  Peter Stone serves as the Executive Director of Sony AI America and receives financial compensation for this work.  The terms of this arrangement have been reviewed and approved by the University of Texas at Austin in accordance with its policy on objectivity in research.}

\bibliographystyle{IEEEtran}
\bibliography{mybib}

\end{document}